\pdfoutput=1
\documentclass[11pt]{article}

\usepackage{authblk}
\usepackage{acl}
\usepackage{times}
\usepackage{latexsym}
\usepackage[T1]{fontenc}
\usepackage[utf8]{inputenc}
\usepackage{microtype}
\usepackage{inconsolata}
\usepackage{graphicx}
\usepackage{smartdiagram}
\usesmartdiagramlibrary{additions}
\usepackage{tabto}
\usepackage{multirow}
\usepackage{tikz}
\usepackage{pgfplots}
\pgfplotsset{compat=1.18}
\usetikzlibrary{patterns}
\usepgfplotslibrary{statistics}

\title{cantnlp@LT-EDI-2024: Automatic Detection of Anti-LGBTQ+ Hate Speech in Under-resourced Languages}

\author[1,2]{Sidney G.-J. Wong}
\author[1]{Matthew Durward}
\affil[1]{University of Canterbury, New Zealand}
\affil[2]{Geospatial Research Institute, New Zealand}
\affil[ ]{\texttt{\{sidney.wong,matthew.durward\}@pg.canterbury.ac.nz}}

\begin{document}
\maketitle

\begin{abstract}
    This paper describes our homophobia/transphobia in social media comments detection system developed as part of the shared task at LT-EDI-2024. We took a transformer-based approach to develop our multiclass classification model for ten language conditions (English, Spanish, Gujarati, Hindi, Kannada, Malayalam, Marathi, Tamil, Tulu, and Telugu). We introduced synthetic and organic instances of script-switched language data during domain adaptation to mirror the linguistic realities of social media language as seen in the labelled training data. Our system ranked second for Gujarati and Telugu with varying levels of performance for other language conditions. The results suggest incorporating elements of paralinguistic behaviour such as script-switching may improve the performance of language detection systems especially in the cases of under-resourced languages conditions.
\end{abstract}

\section{Introduction}
\label{sec:introduction}
    
    The purpose of this shared task was to develop a multiclass classification system to predict instances of homophobia/transphobia in social media comments across different language conditions \cite{kumaresan_overview_2024}. The ten language conditions were: English (\textsc{eng}), Spanish (\textsc{esp}), Gujarati (\textsc{guj}), Hindi (\textsc{hin}), Kannada (\textsc{kan}), Malayalam (\textsc{mal}), Marathi (\textsc{mar}), Tamil (\textsc{tam}), Tulu (\textsc{tcy}), and Telugu (\textsc{tel}).

    The main contribution of this paper is that we extend on the work using spatio-temporally retrained transformer-based language models in \citet{wong_cantnlplt-edi-2023_2023}. We have expanded on the synthetic script-switching approach by incorporating real-world (or organic) samples of script-switching during domain adaptation in the development of our multiclass classification model using pretrained language models.

    \begin{figure}[ht]
        \begin{tikzpicture}[scale=0.9]
            \begin{axis}[
                  ybar,
                  bar width=10pt,
                  ylabel={Observations (000s)},
                  xlabel={Language Condition},
                  x tick label style={rotate=90},
                  symbolic x coords={\textsc{kan},\textsc{tel},\textsc{guj},\textsc{mal},\textsc{mar},\textsc{eng},\textsc{tam},\textsc{hin},\textsc{esp},\textsc{tcy}},
                  xtick=data,
                  ymajorgrids=true,
                  grid style=dashed,
                ]
                  \addplot 
                  [pattern = north west lines]
                  coordinates {
                    (\textsc{kan},12.220)
                    (\textsc{tel},10.990)
                    (\textsc{guj},9.859)
                    (\textsc{mal},4.327)
                    (\textsc{mar},4.250)
                    (\textsc{eng},3.956)
                    (\textsc{tam},3.328)
                    (\textsc{hin},2.880)
                    (\textsc{esp},1.586)
                    (\textsc{tcy},0.73)
                  };
            \end{axis}
        \end{tikzpicture}
        \caption{\label{fig:corpus} Barplot of labelled training data. The combined total number of observations (in thousands) by language condition ordered from the most (\textsc{kan}) to the least (\textsc{tcy}) number of observations.}
    \end{figure}
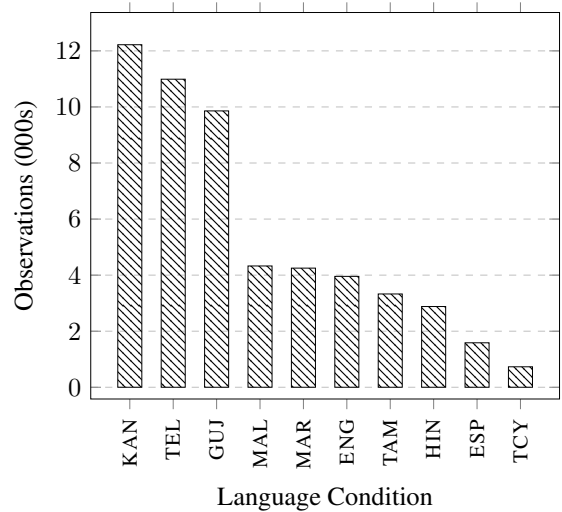

\subsection{Problem Description}
\label{subsec:problemdescription}

    The organisers of the shared task provided labelled training data for each of the ten language conditions. Five of the language conditions belong to the Indo-European language family (\textsc{eng}, \textsc{esp}, \textsc{guj}, \textsc{hin}, and \textsc{mar}) and the remaining five language conditions belong to the Dravidian language family (\textsc{kan}, \textsc{mal}, \textsc{tam}, \textsc{tcy}, and \textsc{tel}).
    
    The labelled training data comes from different sources (\textsc{eng} and \textsc{tam} in \citealp{chakravarthi_dataset_2021}; \textsc{hin} and \textsc{mal} in \citealp{kumaresan_homophobia_2023}; and \textsc{esp} in \citealp{garcia-diaz_umucorpusclassifier_2020}). The training data is made up of comments from users reacting to LGBTQ+ related content on YouTube. The labelled training data for \textsc{guj}, \textsc{kan}, \textsc{mar}, \textsc{tcy}, and \textsc{tel} were introduced for the current shared task.

    The total number of social media comments for each language condition (combining the train and development sets) are shown in Figure \ref{fig:corpus}. \textsc{kan} has the most observations, followed by \textsc{tel} and \textsc{guj}. \textsc{tcy} has the least number of observations. The remaining language conditions each have between 1,000 to 5,000 observations.
    
    The social media comments were manually annotated and broadly labelled using on a three-class classification system \cite{chakravarthi_dataset_2021}. There were only two classes for \textsc{tcy} which we have labelled \textsc{none} and \textsc{homo} for consistency with other language conditions. The classes are:

    \begin{itemize}
        \item \textit{Homophobic Content} (\textsc{homo}): any comments which were deemed gender-based and involved pejorative or defamatory language directed towards non-heterosexual people.
        \item \textit{Transphobic Content} (\textsc{trans}): any derogatory or offensive language directed towards transgender and gender diverse people.
        \item \textit{Non-anti-LGBTQ+ Content} (\textsc{none}): counter speech or hope speech as well as comments which does not contain any homophobic or transphobic content.
    \end{itemize} 

    \begin{table}
        \centering
        \begin{tabular}{lccc}
            \hline
             & \textsc{none} & \textsc{homo} & \textsc{trans}\\
            \hline
            \textsc{eng} & 0.94 & 0.06 & 0.00 \\
            \textsc{esp} & 0.57 & 0.22 & 0.22 \\
            \textsc{guj} & 0.47 & 0.28 & 0.25 \\
            \textsc{hin} & 0.95 & 0.02 & 0.04 \\
            \textsc{kan} & 0.44 & 0.27 & 0.28 \\
            \textsc{mal} & 0.79 & 0.16 & 0.06 \\
            \textsc{mar} & 0.73 & 0.16 & 0.11 \\
            \textsc{tam} & 0.77 & 0.17 & 0.06 \\
            \textsc{tcy} & 0.74 & 0.26 & -  \\
            \textsc{tel} & 0.39 & 0.32 & 0.30 \\
            \hline
        \end{tabular}
        \caption{\label{tab:class} Class distribution by language condition. Note that \textsc{tcy} has a binary class distribution.}
    \end{table}

    The class distribution for each language condition is shown in Table \ref{tab:class}. We observe significant class imbalance between language conditions especially in \textsc{eng} and \textsc{hin} where the \textsc{homo} and \textsc{trans} classes make up less than a tenth of the labelled training data. Of the 3,726 observations in the \textsc{eng} language condition, there are only 221 tokens of \textsc{homo} and nine tokens of \textsc{trans}.
    
    Outside the labelled training data and published material, the organisers did not provide additional corpus or demographic information of the labelled training data as part of the shared task. Therefore, the classification system needs to account for the differences in data availability as well as class imbalance for each language condition.

\subsection{Related Work}
\label{subsec:relatedwork}

    The current shared task is the third shared task on homophobia and transphobia detection in social media comments. The first shared task involved only three language conditions: \textsc{tam}, \textsc{eng}, and a separate \textsc{tam}-\textsc{eng} code-mixed condition \cite{chakravarthi_overview_2022}.

    The classification system with the best performance for \textsc{eng} had a weighted Macro $F_{1}$ score of 0.92 was developed by team \textsc{ablimet} \cite{maimaitituoheti_ablimet_2022} and for \textsc{tam} was 0.94 developed by team \textsc{arguably}. The best performing classification system for the \textsc{tam}-\textsc{eng} code-mixed condition was also developed by team \textsc{arguably} with a weighted Macro $F_{1}$ score of 0.89. The code-mixed condition had the lowest performance across the three conditions.

    Participants took different approaches involving statistical language models and machine learning. The best performing system used \textsc{xlm-roberta} pretrained language models \cite{conneau_unsupervised_2020}. This BERT-based transformer language approach structures the relationship between words with language embeddings \cite{devlin_bert_2019}. These language embeddings account for structures across multilingual conditions.

    The second shared task expanded to five language conditions (\textsc{eng}, \textsc{esp}, \textsc{hin}, \textsc{mal}, and \textsc{tam}) which was broken down by a three-class classification system similar to the current shared task \cite{chakravarthi_overview_2023}. Three of the language conditions (\textsc{eng}, \textsc{mal}, and \textsc{tam}) were further classified into a seven-class classification system.
    
    The weighted Macro $F_{1}$ score for the best performing three-class classification systems was 0.97 for \textsc{eng} and 0.98 for \textsc{hin} developed by \textsc{teamplusone} using BERT-based transformer models. A weight-space ensembling technique presented itself as the best solution for \textsc{esp}, \textsc{mal}, and \textsc{tam} language conditions \cite{ninalga_cordycepslt-edi_2023}.

    The best performing systems for the seven-class classification condition were all developed using transformer language models. The weighted Macro $F_{1}$ score \textsc{eng} was 0.82 developed by team \textsc{teamplusone}, for \textsc{mal} was 0.88 developed by team \textsc{cantnlp} \cite{wong_cantnlplt-edi-2023_2023}, and for \textsc{tam} was 0.87 developed by team \textsc{deepblueai}.
    
    This suggests BERT-based models, such as \textsc{xlm-roberta} for zero-shot learning, are particularly effective in carrying out multiclass classification tasks outlined in the current shared task. More importantly, these systems are simple to implement and allow for domain adaptation \cite{liu_roberta_2019}

    \citet{wong_cantnlplt-edi-2023_2023} introduced synthetically script-switched instances of social media data during domain adaptation to account for the high frequency of script-switching in the labelled data for \textsc{hin}, \textsc{mal}, and \textsc{tam}. The introduction of script-switched language data improved the performance of the homophobia/transphobia detection model in \textsc{hin}, but not \textsc{mal} or \textsc{tam}. 

    The results from \citet{wong_cantnlplt-edi-2023_2023} suggest that there is potential for incorporating paralinguistic behaviour such as script-switching in the development of multiclass detection language systems. Therefore, this paper explores this further by incorporating different forms of script-switching.
    
\begin{figure}
        \begin{tikzpicture}[scale=0.9]
            \begin{axis}
                [
                ytick={1,2,3,4,5,6,7,8,9,10},
                yticklabels={\textsc{guj}, \textsc{kan}, \textsc{tam}, \textsc{tel}, \textsc{mar}, 
                \textsc{mal},\textsc{eng}, \textsc{esp}, \textsc{hin}, \textsc{tcy}},
                ylabel={Language Condition},
                xlabel={Proportion of words in Latin Script(\%)},
                ]
                \addplot+[
                color = black,
                densely dashed,
                boxplot prepared={
                  median=0,
                  upper quartile=0,
                  lower quartile=0,
                  upper whisker=0.7,
                  lower whisker=0,
                },
                ] coordinates {};
                \addplot+[
                color = black,
                densely dashed,
                boxplot prepared={
                  median=0,
                  upper quartile=0,
                  lower quartile=0,
                  upper whisker=0.666667,
                  lower whisker=0,
                },
                ] coordinates {};
                \addplot+[
                color = black,
                densely dashed,
                boxplot prepared={
                  median=0,
                  upper quartile=0.076923,
                  lower quartile=0,
                  upper whisker=0.75,
                  lower whisker=0,
                },
                ] coordinates {};
                \addplot+[
                color = black,
                densely dashed,
                boxplot prepared={
                  median=0,
                  upper quartile=0,
                  lower quartile=0,
                  upper whisker=0.666667,
                  lower whisker=0,
                },
                ] coordinates {};
                \addplot+[
                color = black,
                densely dashdotted,
                boxplot prepared={
                  median=0.166667,
                  upper quartile=1,
                  lower quartile=0,
                  upper whisker=1,
                  lower whisker=0,
                },
                ] coordinates {};
                \addplot+[
                color = black,
                densely dashdotted,
                boxplot prepared={
                  median=0.666667,
                  upper quartile=1,
                  lower quartile=0,
                  upper whisker=1,
                  lower whisker=0,
                },
                ] coordinates {};
                \addplot+[
                color = black,
                densely dotted,
                boxplot prepared={
                  median=1,
                  upper quartile=1,
                  lower quartile=1,
                  upper whisker=1,
                  lower whisker=0.666667,
                },
                ] coordinates {};
                \addplot+[
                color = black,
                densely dotted,
                boxplot prepared={
                  median=1,
                  upper quartile=1,
                  lower quartile=1,
                  upper whisker=1,
                  lower whisker=0.440000,
                },
                ] coordinates {};
                \addplot+[
                color = black,
                densely dotted,
                boxplot prepared={
                  median=1,
                  upper quartile=1,
                  lower quartile=1,
                  upper whisker=1,
                  lower whisker=0,
                },
                ] coordinates {};
                \addplot+[
                color = black,
                densely dotted,
                boxplot prepared={
                  median=1,
                  upper quartile=1,
                  lower quartile=1,
                  upper whisker=1,
                  lower whisker=0,
                },
                ] coordinates {};
              \end{axis}
            \end{tikzpicture}
        \caption{\label{fig:scriptswitch} Boxplot of labelled training data. Language condition by the proportion of observations with at least one word written in Latin script ordered from the lowest (\textsc{tcy}) to the highest (\textsc{guj}) proportion of observations.}
    \end{figure}
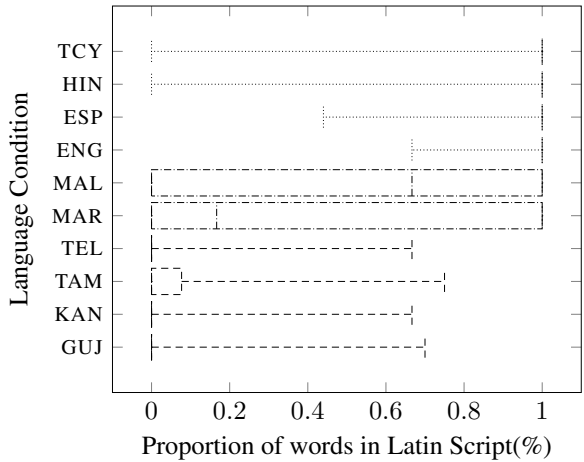
    
\section{Methodology}
\label{sec:methodology}
  
    In this section, we provide an overview of our system development methodology. We took a transformer-based language model approach to develop our system. We used \textsc{xlm-roberta} as the base \textsc{plm} for our system \cite{conneau_unsupervised_2020}. The embeddings in \textsc{xlm-roberta} were trained on two terabytes of web-crawled data for over 100 language including nine of the ten language conditions of interest (with the exclusion of \textsc{tcy}). 
    
    A significant advantage of transformer-based \textsc{plm}s is the ability for domain adaptation as discussed in Section \ref{subsec:domain}. This means we can retrain the default language embedding models with additional language data without the need to train resource-intensive \textsc{plm}s from scratch.

    We tested different forms of script-switching in order to understand the impacts of script-switching on our classification system. We then used the \textsc{plm}s developed in Section \ref{subsec:domain} to fine-tune our multiclass classification model as discussed in Section \ref{subsec:model}. Based on the weighted Macro $F_{1}$ for each language condition, we submitted the results from the best performing multiclass classification system to the organisers.
    
\subsection{Domain Adaptation}
\label{subsec:domain} 

    \begin{table*}
        \centering
        \begin{tabular}{lcccccc}
            \hline
            \multirow{2}{*}{} &
          \multicolumn{2}{c}{\textsc{baseline}} &
          \multicolumn{2}{c}{\textsc{synthetic}} &
          \multicolumn{2}{c}{\textsc{organic}} \\
          & \textit{mono} & \textit{multi} & \textit{mono} & \textit{multi} & \textit{mono} & \textit{multi} \\
            \hline
            \textsc{eng} & 0.32 & \textbf{0.35} & 0.32 & 0.32 & 0.32 & 0.32 \\
            \textsc{esp} & 0.80 & 0.82 & \textbf{0.87} & 0.84 & 0.76 & 0.82 \\
            \textsc{guj} & 0.94 & 0.94 & \textbf{0.95} & \textbf{0.95} & \textbf{0.95} & \textbf{0.95} \\
            \textsc{hin} & \textbf{0.32} & \textbf{0.32} & \textbf{0.32} & \textbf{0.32} & \textbf{0.32} & \textbf{0.32} \\
            \textsc{kan} & 0.92 & 0.93 & \textbf{0.94} & \textbf{0.94} & \textbf{0.94} & \textbf{0.94} \\
            \textsc{mal} & 0.51 & 0.53 & 0.73 & 0.58 & \textbf{0.78} & 0.61 \\
            \textsc{mar} & 0.44 & 0.45 & 0.44 & 0.41 & 0.42 & \textbf{0.46} \\
            \textsc{tam} & 0.48 & 0.42 & 0.54 & 0.49 & 0.48 & \textsc{0.56} \\
            \textsc{tcy} & \textbf{0.72} & 0.43 & 0.43 & 0.43 & 0.43 & 0.43 \\
            \textsc{tel} & \textbf{0.98} & 0.97 & \textbf{0.98} & 0.97 & 0.97 & \textbf{0.98} \\
            \hline
        \end{tabular}
        \caption{\label{tab:performance} Model performance of candidate classification models by Macro $F_{1}$ using our test set split from combining the train and validation sets provided to us by the organisers for each language condition. The three candidate languages models are: \textsc{baseline}, \textsc{synthetic}, and \textsc{organic}. We also compared the performance of language-specific (\textit{mono}) and multilingual (\textit{multi}) multiclass classification models. The best performing system is highlighted in \textbf{bold}.}
    \end{table*}

    The first stage in developing our system involved domain adaptation (also known as retraining). \citet{liu_roberta_2019} noted that domain adaptation can improve the performance of transformer-based language models in downstream tasks. We can do this by introducing domain (or register) specific text samples to produce customised retrained PLMs (or retrained language models). This means we can introduce language data from under-resourced languages such as \textsc{tcy} as well as additional linguistic information such as script-switching - a common phenomenon in social media language. 

    As noted in \citet{wong_cantnlplt-edi-2023_2023}, we observed varying levels of script-switching in the labelled training data. Therefore, we first needed to identify the level of script-switching between language conditions. For each observation, we calculated the proportion of words written in Latin script using the alphabet-detector\footnote{https://pypi.org/project/alphabet-detector/} Python package. 
    
    The proportion of script-switching between language conditions is shown in Figure \ref{fig:scriptswitch} where 0 suggests low usage of Latin-based characters (in the case of \textsc{guj}, \textsc{kan}, \textsc{tam}, \textsc{tel}) while 1 suggests high usage as expected for \textsc{eng} and \textsc{esp}. Figure \ref{fig:scriptswitch} confirms that there is sufficient need to account for the varying-degrees of script-switching between language conditions.
    
    We retrained \textsc{xlm-roberta} with two forms of script-switching: synthetic and organic script-switching. These are our candidate models. We describe how we produced the language data for domain adaptation in Section \ref{subsubsubsec:synthetic} and Section \ref{subsubsubsec:organic}. We produced the candidate language models by retraining the language embeddings using the simpletransformers\footnote{https://simpletransformers.ai/} Python library. We did this over four iterations and we evaluated the training for every 500 steps with AdamW optimisation \cite{loshchilov_decoupled_2019}. Model performance was based on evaluation loss.

\subsubsection{Synthetic Script-Switching}
\label{subsubsubsec:synthetic}

    We took a similar approach as \citet{wong_cantnlplt-edi-2023_2023} to produce synthetic samples of script-switched language data for domain adaptation. We define synthetic as machine-generated texts. Due to the limited availability of observations for some language conditions, our main source of human-generated texts come from the Leipzig Corpus Collection \cite{goldhahn_building_2012}. Each corpus contained 10,000 Wikipedia abstracts produced in 2016 with the exception of \textsc{tcy} which was produced in 2018. 
    
    We then randomly sampled half of the abstracts from each language condition (excluding the Latin-based \textsc{eng} and \textsc{esp}) and used the ai4bharat\footnote{https://pypi.org/project/ai4bharat-transliteration/} Python library to transliterate the relevant Brahmic orthographies into Latin script. Once we produced a subset of synthetically script-switched Wikipedia abstracts, we combined the original abstracts with the synthetically script-switched abstracts. Finally, we combined the labelled training data to create train and evaluation sets. The inclusion of the labelled training data is to ensure register-specific domain adaptation.

\subsubsection{Organic Script-Switching}
\label{subsubsubsec:organic}

    The second form of script-switched language data for domain adaptation involve organic samples of script-switched language data. We define organic as human-generated texts. This proved to be a challenge as we were unable to identify sources of script-switched social media language data for some of the under-resourced language conditions. 
    
    We used the pre-existing labelled training data to produce language profiles to develop a language identification model with the langdetect\footnote{https://pypi.org/project/langdetect/} These language profiles were used to detect organic instances of script-switched social media data from the Global Corpus of Language Use (\textsc{cglu}; \citealp{dunn_mapping_2020}). This produced a train set with 230,000 observations and an evaluation set with 12,000 observations which we could use for domain adaptation.

\subsection{Classification Model}
\label{subsec:model}

    As discussed in Section \ref{subsec:domain}, we developed our multiclass classification models using the candidate language models during the domain adaptation phase. The three candidate languages models are: the baseline \textsc{xlm-roberta} language model (\textsc{baseline}), \textsc{xlm-roberta} retrained with synthetic samples of script-switched language data (\textsc{synthetic}), and \textsc{xlm-roberta} retrained with organic samples of script-switched language data (\textsc{organic}).
    
    We resampled the available data to create our own train (80\%), validation (10\%), and test (10\%) sets to avoid over-fitting on the validation set during model evaluation. We trained language specific classification models (\textit{mono}) and an ensemble multilingual classification model (\textit{multi}) by combining the labelled training data. We trained the multiclass classification model using the simpletransformers Python package for four iterations and we evaluated the training for every 500 steps with AdamW optimisation \cite{loshchilov_decoupled_2019}. The model performance was based on evaluation loss.

    The model performance for each of the candidate models are shown in Table \ref{tab:performance}. We have indicated the best performing model based on average Macro $F_{1}$ score (highlighted in \textbf{bold}). In some language conditions, there were multiple best performing models. Not included in Table \ref{tab:performance} are the combined average Macro $F_{1}$ score for the multilingual models: the average Macro $F_{1}$ for the \textsc{baseline} model was 0.89; both \textsc{synthetic} and \textsc{organic} models had an average Macro $F_{1}$ of 0.90.

    \begin{table}
        \centering
        \begin{tabular}{lccc}
            \hline
             & \textsc{cantnlp} & Best Performance\\
            \hline
            \textsc{eng} & 0.323 & \textit{0.496} \\
            \textsc{esp} & 0.496 & \textit{0.582} \\
            \textsc{guj} & 0.962 & \textit{0.968} \\
            \textsc{hin} & 0.326 & \textit{0.458} \\
            \textsc{kan} & 0.943 & \textit{0.948} \\
            \textsc{mal} & 0.775 & \textit{0.942} \\
            \textsc{mar} & 0.433 & \textit{0.626} \\
            \textsc{tam} & 0.555 & \textit{0.880} \\
            \textsc{tcy} & 0.452 & \textit{0.707} \\
            \textsc{tel} & 0.965 & \textit{0.971} \\
            \hline
        \end{tabular}
        \caption{\label{tab:results} The average Macro $F_{1}$ score of our classification system, and the average Macro $F_{1}$ score of the overall best performing classification system.}
    \end{table}

\section{Results}
\label{sec:results}

    Based on the average Macro $F_{1}$ score of the candidate models as shown in Table \ref{tab:performance}, we nominated the language-specific synthetic classification model as the best performing classification system. We applied this classification system and submitted the results to the organisers. The results of our submitted homophobia/transphobia detection system are shown in Table \ref{tab:results}. The best performing language condition was \textsc{tel} with an average Macro $F_{1}$ of 0.97 and the worst performing language condition was \textsc{eng} with an average Macro $F_{1}$ of 0.32.

    Our final rank for each language conditions are as follows: for \textsc{eng} we came tenth equal out of ten teams;
    for \textsc{esp} we came fourth out of five teams; for \textsc{guj} we came second out of six teams; for \textsc{hin} we came fourth equal out of seven teams; for \textsc{kan} we came fourth out of eight teams; for \textsc{mal} we came seventh out of nine teams; for \textsc{mar} we came fifth out of six teams; for \textsc{tam} we came fifth out of eight teams; for \textsc{tcy} we came third equal out of four teams; and finally for \textsc{tel} we came second out of nine teams.\footnote{Note the final rankings differ from the published results for \textsc{esp}, \textsc{tam}, and \textsc{tel} as they were not included in the final rank list due to human error from the organising committee during submission.}

\section{Discussion}
\label{sec:discussion}

    The use of synthetic and organic script-switched language data during domain adaptation increased the performance for all language conditions from the \textsc{baseline} model with the exception of \textsc{eng}, \textsc{hin}, and \textsc{tcy}. We expected the \textsc{eng} and \textsc{esp} language conditions to perform poorly with our proposed methodology as there were very few instances of script-switching, but the poor performance of \textsc{tcy} was unexpected.

    We hypothesise the poor performance in \textsc{tcy} was due to the limited number of observation (as shown in Figure \ref{fig:corpus}) the higher than expected usage of Latin-based script for \textsc{tcy} in the labelled training data (as shown in Figure \ref{fig:scriptswitch}). This will require robust statistical analysis beyond the scope of the current paper.

    We also posit the poor performance of \textsc{eng} and \textsc{hin} was a result of the class imbalance between instances of homophobic, transphobic and the non-anti-LGBTQ+ content as demonstrated in Table \ref{tab:class}. The performance of our \textsc{eng} and \textsc{hin} language-specific detection models are in line with other participating teams.
    
    In contrast to the method proposed in \citet{wong_cantnlplt-edi-2023_2023}, we did not include any methods to counter the class imbalance in the training data nor did we include random noise injection to expand the minority classes. It was shown that random over sampling of minority classes did not significantly improve the performance of the detection models.

\section{Conclusion}
\label{sec:conclusion}
    
    The main contribution of the current paper is the proposal to use synthetic and organic script-switching examples of during domain adaptation to improve the down-stream performance for under-resourced languages. We demonstrated that our methodology improved the model performance for \textsc{guj}, \textsc{kan}, \textsc{mal}, \textsc{mar}, and \textsc{tam} even though the improvement was only marginal. Even though our homophobia/transphobia detection system did not rank first for any of the ten language conditions, we were pleased with the performance of our detection system which supports the inclusion of paralinguistic information.

\section*{Ethics Statement}

    The purpose of the current shared task is to develop a homophobic/transphobic language detection system in social media texts particularly for under-resourced Indo-Aryan and Dravidian languages within the fields of computational linguistics and natural language processing.
    
    We recognise the importance of community-lead research in particular by members of under-represented and minoritised communities. The lead author acknowledges his positionality as a member of the LGBTQ+ community. The lead author is familiar with anti-LGBTQ+ discourse both in online and offline spaces and the harmful effects of hate speech and offensive language on members of the LGBTQ+ communities \cite{wong_queer_2023}.
    
    In terms of the authors' linguistic membership, the authors share proficiency in \textsc{eng} and \textsc{esp}; however, the authors acknowledge their limited experience with \textsc{guj}, \textsc{kan}, \textsc{mar}, \textsc{tam}, \textsc{tcy}, and \textsc{tel} with some exposure to \textsc{hin} and \textsc{mal}. We acknowledge the limitations of our analysis in language conditions where we have limited proficiency and we will follow the guidance and expertise of members from the relevant language communities.
    
    We want to thank the organisers of the shared task and the workshop on Language Technology for Equality, Diversity, and Inclusion. We also want to thank the contributors of the training data and those who were involved in the labelling process across the different language conditions.

\section*{Limitations}
    
    Under the purview of developing a homophobic/transphobic language detection system in social media texts, we want to highlight the limitations of our proposed system and methodology.
    
    Firstly, we acknowledge there are differences in data quality and veracity between the different language conditions. This is based on the differences in the corpus size between the different language conditions (as shown in Figure \ref{fig:corpus}) and the distribution of homophobic and transphobic content.

    In light of these data quality issues, we have not accounted for these differences between language conditions. This means we do not entirely understand the downstream impacts on model performance - although it is clear that there is a possible relationship between larger and more balanced language conditions (\textsc{tel}) performing better than smaller and more imbalanced language conditions (\textsc{tcy}). It is possible these differences could exacerbate biases already observed in transformer-based language models \cite{bhardwaj_investigating_2021}.

    Beyond the upstream and downstream impacts of bias in transformer-based language models, we also recognised that incorporating external data sets from \textsc{lcc} \cite{goldhahn_building_2012} and the \textsc{cglu} \cite{dunn_mapping_2020} introduces additional biases not properly addressed in this paper such as geographic bias in social media language data \cite{wong_comparing_2022}.

    Secondly, there is a need to conduct this form of research under a sociolinguistic or linguistic anthropological framework. There is a risk that training data detecting homophobia, transphobia, hate speech, or offensive may not necessarily reflect the social, political, or linguistic realities of different populations. This is because some of the features extracted from the labelled training data may not reflect real-world knowledge.
    
    These differences are particularly evident when we apply these detection systems across dialect contexts \cite{wong_monitoring_2023}. For this reason, we propose that future work in this area should also consider how these systems perform in real-world context beyond the evaluation of labelled training data. We should work alongside members of LGBTQ+ communities from culturally and linguistically diverse backgrounds to understand the effectiveness and generalisability of our homophobic/transphobic detection systems.

\bibliography{references}

\end{document}